\documentclass{article}

\usepackage{arxiv}

\usepackage[utf8]{inputenc} 
\usepackage[T1]{fontenc}    
\usepackage{hyperref}       
\usepackage{url}            
\usepackage{booktabs}       
\usepackage{amsfonts}       
\usepackage{nicefrac}       
\usepackage{microtype}      
\usepackage{lipsum}
\newcommand{\rankfirst}[1]{\textbf{#1}}
\newcommand{\ranksecond}[1]{#1}
\newcommand{\rankthird}[1]{#1}
\usepackage{graphicx}
\graphicspath{ {./images/} }

\usepackage[table]{xcolor}
\usepackage{graphicx}
\usepackage{capt-of}
\usepackage{multirow}
\usepackage{subcaption}
\usepackage{amsmath}

\title{DuoWM: Integrating Dual 6-DoF Visual Flows into
Action-Conditioned Drone World Model}

\author{%
  Zixuan Li\textsuperscript{1,\dag},
  Kangyao Huang\textsuperscript{2},
  Meng Xu\textsuperscript{3},
  Wan Tian\textsuperscript{4} \\
  \bfseries
  Xiaoyu Tang\textsuperscript{5},
  Jinge Sun\textsuperscript{6},
  Yingliang Yao\textsuperscript{7},
  He Chen\textsuperscript{8} \\
  \bfseries
  Lingzhong Meng\textsuperscript{9},
  Yunzhi Xue\textsuperscript{10},
  Jingyu Chen\textsuperscript{11,*} \\[0.45em]
  \normalfont\normalsize
  \textsuperscript{1}University of International Business and Economics (uibe.edu.cn) \\
  \normalfont\normalsize
  \textsuperscript{2}Tsinghua University (tsinghua.edu.cn) \\
  \normalfont\normalsize
  \textsuperscript{3}University of International Business and Economics (uibe.edu.cn) \\
  \normalfont\normalsize
  \textsuperscript{4}Peking University (pku.edu.cn) \\
  \normalfont\normalsize
  \textsuperscript{5}University of the Chinese Academy of Sciences (ucas.ac.cn) \\
  \normalfont\normalsize
  \textsuperscript{6}QiYuanLab (qiyuanlab.com) \\
  \normalfont\normalsize
  \textsuperscript{7}University of International Business and Economics (uibe.edu.cn) \\
  \normalfont\normalsize
  \textsuperscript{8}Beihang University (buaa.edu.cn) \\
  \normalfont\normalsize
  \textsuperscript{9}Institute of Software, Chinese Academy of Sciences (iscas.ac.cn) \\
  \normalfont\normalsize
  \textsuperscript{10}Institute of AI for Industries (iaii.ac.cn) \\
  \normalfont\normalsize
  \textsuperscript{11}Institute of Software, Chinese Academy of Sciences (iscas.ac.cn) \\[0.35em]
  \normalfont\normalsize
  \textsuperscript{\dag}First author. \quad
  \textsuperscript{*}Corresponding author.
}

\begin{document}
\maketitle
\begin{abstract}
Action-conditioned video world models generate future observations from an initial frame and a camera trajectory, 
but often fail to align camera commands with visual dynamics or represent full 6-DoF scene transformations.
We present DuoWM, a first-person drone video world model with two complementary visual conditions.
Action flow projects future camera positions and orientations into each target frame, while Object flow forward-warps the initial observation 
and overlays local 6-DoF transformations to ground the induced scene motion.
We further introduce transform-aware pretraining, which pairs object appearance with explicit transformation axes before standard RGB video training.
On Sekai-drone dataset, DuoWM achieves the best results across video quality, 3D consistency, and trajectory following.
Compared with Captain Safari, the state-of-the-art method, our approach reduces real-domain FVD and LPIPS by 17.2\% and 21.6\%, respectively, 
while improving game-domain AUC@30 from 0.5003 to 0.5719 and AUC@15 from 0.3088 to 0.3747. 
\end{abstract}


\section{Introduction}
\label{sec:intro}
Interactive world models are emerging as a promising training paradigm for embodied agents, enabling them to predict visual observations under different action commands~\cite{wu2024ivideogpt,yu2025wonderworld}. Recent advances in aerial robotics have demonstrated increasingly capable navigation and transportation policies~\cite{chen2023deep,huang2021decentralised,chen2023behavioural,chen2025meta}, motivating high-fidelity training environments that capture the visual consequences of complex flight maneuvers. While physics-based simulators rely on manually engineered dynamics and environments~\cite{huang2026airgym,xu2025navrl}, video world models offer a data-driven alternative by learning visual dynamics from videos. Recent aerial world models~\cite{zhao2025airscape,serpiva2026flightdiffusion} demonstrate the potential of this approach, yet maintaining correspondence between commanded camera trajectories and generated visual dynamics remains challenging.

The core challenge lies in establishing a reliable correspondence between motion conditions and their visual consequences. 
Existing approaches explore several forms of motion conditioning, including learned latent actions, image-space controls, and motion representations based on point trajectories of the objects, optical flow, and 3D bounding boxes \cite{geng2025motion,wang2025ati,bharadhwaj2024track2act,li2026novaflow,wang2024boximator}. 
However, these approaches typically lack precise frame-wise control over evolving video sequences, often relying on initial-frame guidance propagated across subsequent frames.
Moreover, their motion representations primarily capture 3-DoF translations \cite{harley2025alltracker,karaev2025cotracker3}, leaving full 6-DoF (six-degree-of-freedom) rigid transformations insufficiently modeled. 

Explicit pose visualization and geometry-aware training already provide useful precedents: EnerVerse-AC renders end-effector positions and orientation axes as visual action maps~\cite{jiang2025enerverse}, while PoseTraj first reconstructs videos with projected 3D bounding boxes and subsequently removes box supervision to refine appearance~\cite{ji2025posetraj}. These advances raise a key question for aerial video generation: how can frame-aligned camera commands and scene transformation priors be jointly represented, and how can training align their geometric correspondence with visual appearance?

We address this question with DuoWM, an action-conditioned drone world model using \emph{dual 6-DoF visual flows}. \emph{Action flow} renders target camera positions and orientation axes to align commanded trajectories with target frames. \emph{Object flow} combines first-frame forward warping with local transformation axes derived using MoSE3~\cite{cheng2026mose3}. MoSE3 supplies dense per-pixel $\mathrm{SE}(3)$ estimates; DuoWM uses these estimates to construct visual conditions for video generation. We further use \emph{transform-aware pretraining} to connect these visual conditions with reconstruction targets. 
The model first reconstructs RGB frames overlaid with local 6-DoF transformation axes, then switches to unmarked RGB targets while both flows remain as conditions.
This representation differs from PoseTraj~\cite{ji2025posetraj}, which uses projected object boxes only as temporary supervision and then removes them. 
Our 6-DoF axes encode local rigid motion and are fed to the generator through \emph{Action flow} and \emph{Object flow}.

We evaluate DuoWM along three axes: \emph{video quality}, \emph{3D consistency}, and \emph{trajectory following}.
Across these criteria, our method consistently outperforms prior camera-controlled video generators on Sekai-game-drone and Sekai-real-drone.
Table~\ref{tab:main} and Table~\ref{tab:real_results} report gains in perceptual quality, geometric consistency, and trajectory accuracy under complex flight maneuvers.
Compared with Captain Safari, DuoWM reduces real-domain FVD and LPIPS by 17.2\% and 21.6\%, and improves game-domain AUC@30 and AUC@15 by 14.3\% and 21.3\%.
Qualitative comparisons (Figure.~\ref{fig:method_comparison} and Figure.~\ref{fig:3d_construction}) further show stable scene geometry and closer adherence to the commanded camera trajectory.

In summary, our contributions are:
\begin{itemize}
    \item We develop a dual-flow conditioning framework for aerial video generation, combining frame-aligned camera commands with local $\mathrm{SE}(3)$ cues derived from first-frame-based warped observations.
    \item We propose 6-DoF transform-aware pretraining to establish the geometric correspondence between object appearances and explicit 6-DoF transformation axes, connecting transform-aware pretraining with the visual conditions used during RGB generation.
    \item Experimental results show that DuoWM improves visual fidelity and camera trajectory following on the Sekai benchmark, 
    demonstrating the effectiveness of dual-flow conditioning and 6-DoF pretraining for action-conditioned drone video generation.
\end{itemize}

\section{Related Work}
\subsection{Action-Conditioned Video Generation and World Models}
Latent diffusion models and diffusion transformers have substantially advanced high-fidelity image and video synthesis~\cite{rombach2021highresolution,peebles2023scalable}. Building on these foundations, large-scale video generators such as Sora~\cite{openai2024sora} and Wan~\cite{wan2025wan} demonstrate increasingly realistic long-range generation, while Genie~\cite{bruce2024genie} and iVideoGPT~\cite{wu2024ivideogpt} extend generation toward environments that respond to user actions. Autoregressive denoising further carries action-conditioned interaction across longer temporal contexts~\cite{zhu2026astra}. These interactive generators are now used as world models for embodied navigation~\cite{bar2025navigation} and visuomotor control and planning~\cite{kim2026cosmos}.

Existing approaches condition video synthesis on explicit camera trajectories~\cite{he2024cameractrl}, disentangled camera and object motion~\cite{wang2024motionctrl,shi2024motion}, or user-specified object trajectories~\cite{geng2025motion}. These controls are further shaped by 3D pose constraints~\cite{ji2025posetraj} and propagated through latent trajectory guidance~\cite{chu2026wan}. Interactive generation additionally combines camera navigation with object manipulation~\cite{duo_worldcraft}. DualCamCtrl~\cite{zhang2026dualcamctrl} adopts two branches to jointly generate RGB and depth for geometry-aware camera control. DuoWM instead retains a pretrained video backbone and supplies two complementary visual conditions: \emph{Action flow} represents the commanded aerial trajectory at each target frame, while \emph{Object flow} combines warped appearance with local $\mathrm{SE}(3)$ transformation cues.

\subsection{Visual Conditioning and 3D Spatial Awareness for World Models}
Visual action representations make control signals spatially compatible with video generators. Existing methods encode motion as dense motion fields~\cite{niu2024mofa}, bounding boxes~\cite{wang2024boximator}, point trajectories~\cite{zhou2025trackgo}, pose trajectories~\cite{ji2025posetraj}, or warped observations~\cite{kong2025worldwarp}. For embodied prediction, point tracks~\cite{bharadhwaj2024track2act} and generated flow~\cite{li2026novaflow} link visual motion to manipulation, while optical flow~\cite{chen2026flowwam} and projected point motion~\cite{li2026hydra} serve as action interfaces for world modeling and control. EnerVerse-AC~\cite{jiang2025enerverse} provides a direct precedent for visual pose encoding: it projects end-effector positions and renders orientation axes, with gripper openness encoded by shading. DuoWM adapts this visual representation to aerial camera trajectories and combines it with scene-transformation conditions. MoSE3~\cite{cheng2026mose3} supplies the dense world-space $\mathrm{SE}(3)$ estimates underlying those local cues; our contribution concerns their use as generative conditions, rather than a new motion estimator.

Complementary work introduces spatial perception for video generation through 4D scene representations~\cite{zhen2025tesseract}, geometry grounding~\cite{yang2026dreamworld}, pose-aligned 3D memory~\cite{chou2026captain}, and occlusion-aware layouts~\cite{agrawal2026seethrough3d}. Other approaches incorporate physics-based interaction~\cite{zhang2024physdreamer}, rigid-body simulation~\cite{liu2024physgen}, physics-informed embodied modeling~\cite{shang2026roboscape}, or physics-reinforced simulation~\cite{zhang2026physisforcing}. DuoWM focuses on image-space conditioning: local transformation cues are represented alongside scene appearance, and transform-aware pretraining prepares the generator to use these conditions before standard RGB training.

\section{Method}
\label{sec:method}
Given an initial RGB image $I_0\in\mathbb R^{H\times W\times3}$, a text prompt $y$, and a target camera trajectory, DuoWM generates a trajectory-conditioned drone video. Here, $H$ and $W$ denote image height and width. The trajectory comprises camera-to-world poses $C_t\in\mathrm{SE}(3)$ and intrinsics $K_t\in\mathbb R^{3\times3}$ at frame indices $t=0,\ldots,T-1$, where $T$ is the video length. The group $\mathrm{SE}(3)$ represents three-dimensional rigid transformations as $4\times4$ matrices.

DuoWM combines two image-space conditions with a pretrained camera-conditioned video diffusion transformer (DiT). \emph{Action flow} expresses future camera positions and orientations relative to each current target view. \emph{Object flow} combines first-frame forward-warped appearance with local geometric axes and tracks. Transform-aware pretraining associates these conditions with appearance through geometry-annotated reconstruction targets, followed by natural RGB video training.

\begin{figure*}[t]
    \centering
    \includegraphics[width=0.9\textwidth]{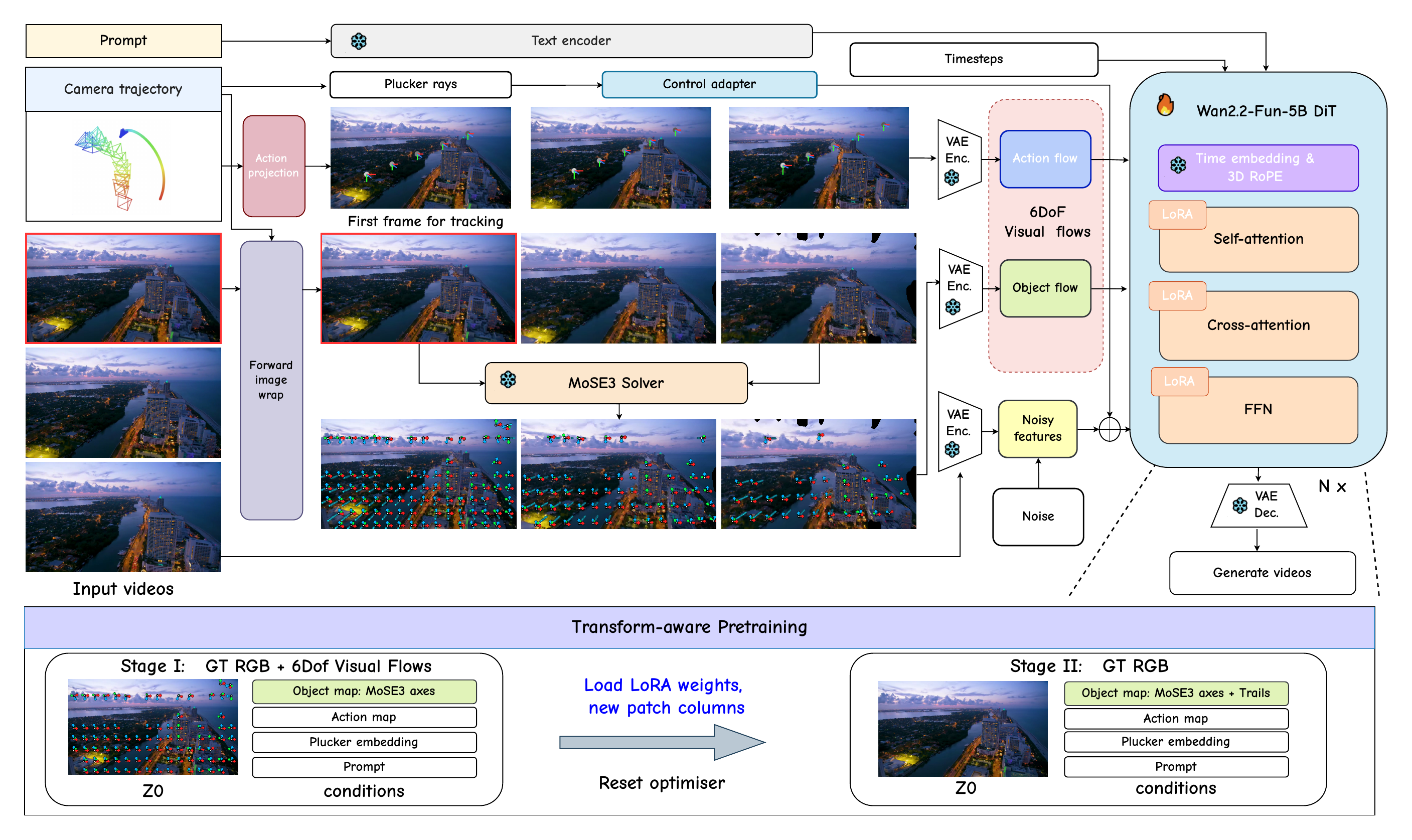}
    \caption{\textbf{Overview of DuoWM.} Action Flow renders camera commands in each target view. Object Flow combines first-frame forward warping with MoSE3-derived local axes and tracks. A frozen video variational autoencoder encodes both flows, which are concatenated with the noisy video and first-frame conditions before patch embedding. Native text cross-attention and the Pl\"ucker-ray camera adapter are retained. Pretraining reconstructs RGB videos overlaid with local axes; subsequent training reconstructs unmarked RGB videos.}
    \label{fig:overview}
\end{figure*}

\subsection{Action Flow}
\label{sec:action_flow}
Inspired by the visual pose encoding of EnerVerse-AC~\cite{jiang2025enerverse}, \emph{Action flow} converts camera motion into frame-aligned position and orientation markers. We resample the trajectory at video timestamps using constant-body-velocity interpolation on $\mathrm{SE}(3)$ and linear interpolation of focal lengths and principal points.

For a future frame index $j>t$, its pose relative to the current camera is
\begin{equation}
C_{t\rightarrow j}=C_t^{-1}C_j=
\begin{bmatrix}R_{t,j}&\mathbf d_{t,j}\\\mathbf0^{\mathsf T}&1\end{bmatrix},
\label{eq:action_relative_pose}
\end{equation}
where $R_{t,j}\in\mathbb R^{3\times3}$ and $\mathbf d_{t,j}\in\mathbb R^3$ are the relative rotation and translation. The reference frame is updated at every timestamp, with its positive optical axis defining the current forward direction.

Let $\ell>0$ denote the rendered axis length and $\mathbf e_i\in\mathbb R^3$ the $i$-th standard basis vector. We represent each future pose by its center and three local endpoints, $\mathbf q\in\{\mathbf0,\ell\mathbf e_1,\ell\mathbf e_2,\ell\mathbf e_3\}$. Their pixel coordinates are
\begin{equation}
\mathbf p_{t,j}(\mathbf q)=\pi\!\left(K_t[\mathbf d_{t,j}+R_{t,j}\mathbf q]\right),
\label{eq:action_projection}
\end{equation}
where $\pi([a,b,c]^{\mathsf T})=(a/c,b/c)^{\mathsf T}$ denotes perspective division. The projected center encodes position, while the three projected axes convey orientation relative to the current camera.

Valid centers must have positive depth and lie inside the image; axis segments are clipped at image boundaries. We render them on a dark RGB map $A_t\in[0,1]^{H\times W\times3}$. Centers have a fixed 50-pixel radius, and their depth $z_{t,j}=\mathbf e_3^{\mathsf T}\mathbf d_{t,j}$ determines an 8-bit gray value $g(z_{t,j})$:
\begin{equation}
\begin{aligned}
\gamma(z_{t,j})&=\operatorname{clip}\!\left(
\frac{\log z_{t,j}-\log z_{\mathrm{near}}}{\log z_{\mathrm{far}}-\log z_{\mathrm{near}}},0,1\right),\\
g(z_{t,j})&=220-140\gamma(z_{t,j}).
\end{aligned}
\label{eq:action_depth}
\end{equation}
Here, $z_{\mathrm{near}}$ and $z_{\mathrm{far}}$ are the 5th and 99.5th percentiles of the reference depths, and $\operatorname{clip}$ bounds its argument to $[0,1]$. Nearer centers are brighter; gray values are divided by 255 for the normalized map. Red, green, and blue segments represent the local $x$, $y$, and $z$ axes. Longer-horizon poses are drawn first to preserve nearer-term commands where markers overlap.

Repeating this process produces \emph{Action flow} $\mathcal A=\{A_t\}_{t=0}^{T-1}$. Frames without valid projected centers contain only the background, including pure rotation about a fixed center. The native Pl\"ucker-ray pathway remains active for all target poses, complementing these visibility-dependent image-space cues.

\subsection{Object Flow}
\label{sec:object_flow}
\emph{Object flow} grounds camera commands in the geometry of the initial observation. It combines forward-warped appearance, local coordinate axes, and visible tracks estimated using frozen MoSE3~\cite{cheng2026mose3}.

\paragraph{Forward-warped appearance.}
MoSE3 predicts initial depth $D_0\in\mathbb R^{H\times W}$ from the first frame. For source pixel $\mathbf p=(u,v)^{\mathsf T}$, define its homogeneous coordinate $\widetilde{\mathbf p}=(u,v,1)^{\mathsf T}$. Let $\bar D_0=s_dD_0$ be the calibrated depth, where $s_d>0$ aligns depth with the camera-translation units of dataset domain. Its projection into target view $t$ is
\begin{equation}
\mathbf p_t=\pi\!\left(
K_t\begin{bmatrix}I_3&\mathbf0\end{bmatrix}C_t^{-1}C_0
\begin{bmatrix}\bar D_0(\mathbf p)K_0^{-1}\widetilde{\mathbf p}\\1\end{bmatrix}\right),
\label{eq:forward_projection}
\end{equation}
where $I_3$ is the $3\times3$ identity matrix. The relative transform changes coordinates without normalizing translation magnitude.

We fit one scale per domain using training videos, minimizing Huber reprojection error against tracked static-background correspondences under known camera poses. Errors are averaged within videos and then across videos, excluding invalid matches, dynamic regions, and clips with insufficient translational parallax. The resulting scale is fixed for caching, training, and inference. This calibration assumes consistent translation units within each domain and does not use test-video correspondences.

We forward-splat source colors and retain the nearest valid point at each destination pixel using a depth buffer. This produces a warped image $W_t\in\mathbb R^{H\times W\times3}$ and a coverage mask $M_t\in\{0,1\}^{H\times W}$. Uncovered regions remain invalid. The sequence reflects the initial scene under the prescribed camera motion, without future RGB observations.

\paragraph{Local geometric markers.}
MoSE3 tracks sampled surface anchors through the warped sequence in its reconstructed coordinate system. For anchor index $k$, let $\mathbf x_{0,k}\in\mathbb R^3$ be its initial position and $F_{0,k}\in\mathbb R^{3\times3}$ its local frame, whose columns are coordinate axes. The estimated local rotation $R^{\mathrm{loc}}_{t,k}\in\mathbb R^{3\times3}$ and translation $\mathbf d^{\mathrm{loc}}_{t,k}\in\mathbb R^3$ yield
\begin{equation}
\begin{aligned}
\mathbf x_{t,k}&=R^{\mathrm{loc}}_{t,k}\mathbf x_{0,k}+\mathbf d^{\mathrm{loc}}_{t,k},\\
F_{t,k}&=R^{\mathrm{loc}}_{t,k}F_{0,k}.
\end{aligned}
\label{eq:local_transform}
\end{equation}
We project these anchors and axes using MoSE3-reconstructed camera-to-world poses $\widehat C_t\in\mathrm{SE}(3)$ and intrinsics $\widehat K_t\in\mathbb R^{3\times3}$. Using reconstructed geometry and cameras together keeps rendering in a common coordinate system, distinct from the prescribed cameras used for warping.

We overlay consistently colored axes and short traces of recently visible anchor positions on $W_t$. Invalid anchors are removed, traces stop at visibility gaps, and marks outside $M_t$ are suppressed. The resulting \emph{object flow} is $\mathcal O=\{O_t\}_{t=0}^{T-1}$, with $O_t\in\mathbb R^{H\times W\times3}$. It encodes viewpoint-dependent local geometric correspondences, rather than observed future independent object motion.

\subsection{Transform-Aware Pretraining}
\label{sec:geometric_warmup}
The generator must connect geometric markers with scene appearance to use the two flows. We establish this correspondence through geometry-annotated reconstruction before training on unmarked RGB videos.

\paragraph{Geometry-annotated reconstruction.}
Let $\mathcal I=\{I_t\}_{t=0}^{T-1}$ denote a training RGB video. Frozen MoSE3 supplies local transformation axes rendered as an RGB layer $\Gamma_t\in\mathbb R^{H\times W\times3}$. Its binary support mask $B_t\in\{0,1\}^{H\times W}$ identifies marker pixels. The pretraining target is
\begin{equation}
I_t^{\mathrm{geo}}=(1-B_t)\odot I_t+B_t\odot\Gamma_t,
\label{eq:warmup_target}
\end{equation}
where $\odot$ denotes element-wise multiplication and the mask is broadcast over RGB channels. The target sequence $\mathcal I^{\mathrm{geo}}=\{I_t^{\mathrm{geo}}\}_{t=0}^{T-1}$ preserves appearance outside the markers, placing local geometry and appearance in the same reconstruction space.

Full-clip information is used only to construct these supervised targets. Both flows are constructed from the clean initial image and prescribed trajectory, and the native first-frame condition remains unmarked. During pretraining, \emph{Object flow} contains warped appearance and local axes without traces.

\paragraph{RGB video training.}
We transfer the pretrained parameters, reset the optimizer, and replace $\mathcal I^{\mathrm{geo}}$ with the unmarked target $\mathcal I$. \emph{Object flow} now includes both axes and traces. The architecture and training objective remain unchanged, while the generated target and \emph{object flow} rendering switch together. This stage trains the generator to use geometric conditions while producing natural RGB appearance.

\subsection{Conditional Generation and Optimization}
\label{sec:conditional_generation}
\paragraph{Condition injection.}
Let $\mathcal E$ denote the frozen video variational autoencoder, or VAE, encoder. We encode both flows and align their latent channel statistics with those of the image-reference latent, obtaining $z_A$ and $z_O$. Spatial resizing and temporal pooling of the coverage masks produce a one-channel mask $m$ on the VAE grid.

Let $z_\sigma$ denote the noisy video latent at noise level $\sigma$, and let $c_0$ denote the native first-frame condition. The initial hidden tokens $H^{(0)}$ are
\begin{equation}
H^{(0)}=P_\theta\!\left([z_\sigma;c_0;z_A;z_O;m]\right)+\mathcal A_{\mathrm{cam}}(r),
\label{eq:dual_flow_input}
\end{equation}
where brackets denote channel-wise concatenation and $P_\theta$ is the expanded patch projection. The native camera adapter $\mathcal A_{\mathrm{cam}}$ maps the target-camera Pl\"ucker-ray representation $r$ to additive token features. A frozen text encoder maps $y$ to text features $c_{\mathrm{txt}}$, supplied through the original cross-attention layers. Thus, both flows enter before Transformer processing, while native camera and text conditioning retain their original pathways.

We train the added patch-projection columns and rank-32 low-rank adaptation, or LoRA, parameters in the attention and feed-forward layers. These trainable parameters are denoted collectively by $\theta$. Original backbone parameters, original projection columns, the camera adapter, the VAE, and the text encoder remain frozen.

\paragraph{Training objective.}
Let $\mathcal J$ denote the target video, equal to $\mathcal I^{\mathrm{geo}}$ during pretraining and $\mathcal I$ during RGB training, and define $z=\mathcal E(\mathcal J)$. Following flow matching~\cite{lipman2023flow,esser2024scaling}, we sample a standard Gaussian tensor $\epsilon$ with the same shape as $z$. A noise level $\sigma\in[0,1]$ is sampled using the backbone's training scheduler. The interpolation and target velocity are
\begin{equation}
z_\sigma=(1-\sigma)z+\sigma\epsilon,
\qquad v^\star=\epsilon-z.
\label{eq:flow_interpolation}
\end{equation}
Let $\mathcal C=\{c_0,z_A,z_O,m,r,c_{\mathrm{txt}}\}$ collect the generation conditions. The denoiser $v_\theta$ predicts velocity under the objective
\begin{equation}
\mathcal L=\mathbb E_{\mathcal J,\sigma,\epsilon}
\left[w(\sigma)\operatorname{MSE}\!\left(v_\theta(z_\sigma,\sigma;\mathcal C),v^\star\right)\right],
\label{eq:training_objective}
\end{equation}
where $\operatorname{MSE}$ averages squared errors over latent elements and $w(\sigma)$ is the scheduler's loss weight. Both training stages use this objective.

\paragraph{Inference.}
We construct and encode the two flows once from the initial image and target cameras. Starting from Gaussian video noise, the backbone sampler integrates the predicted velocity toward the clean latent, which the frozen VAE decoder converts to RGB frames. MoSE3 is used only for condition preparation; neither future RGB observations nor geometry-annotated targets are required at inference.

\section{Experiments}
\label{sec:exp}
\subsection{Experimental Setup}
\paragraph{Datasets and evaluation.}
We evaluate DuoWM on two datasets~\cite{li2025sekai}: Sekai-game-drone, comprising 5,118 training and 474 test videos, and Sekai-real-drone, comprising 4,368 training and 402 test videos. 
Each test sample consists of an initial RGB frame, a text prompt, and a target camera trajectory. 
We use 121-frame video sequences at a resolution of $1280\times704$ and 24 FPS, with camera poses and intrinsics aligned to the corresponding video timestamps. 
We assess video quality using FVD~\cite{unterthiner2018towards} and LPIPS~\cite{zhang2018unreasonable}, 3D consistency using MEt3R~\cite{asim2025met3r}, 
and camera relocation accuracy using AUC@30, AUC@15~\cite{wang2025vggt}, and trajectory cosine similarity (CosSim). 
MEt3R is computed between ground-truth and generated frames at matched timestamps.
\paragraph{Evaluation protocol.}
We follow Captain Safari's evaluation scripts for FVD, LPIPS, and MEt3R.
For trajectory evaluation, we recover camera poses using HLoc and COLMAP.
AUC summarizes relative rotation and translation-direction errors at
$15^\circ$ and $30^\circ$.
CosSim compares flattened camera extrinsics after first-frame
canonicalization, without Sim(3) alignment.
Videos without recovered poses are excluded from trajectory averages.
\paragraph{Baselines.}
We compare DuoWM with Wan2.2-5B base, Wan2.2-Control-Camera, Real-CamI2V~\cite{zheng2024cami2v}, CameraCtrl~\cite{he2024cameractrl}, MotionCtrl~\cite{wang2024motionctrl}, 
and Captain Safari~\cite{chou2026captain}. These baselines cover diverse paradigms of camera-controlled video generation, 
including video generation base model, camera-conditioned generation and memory-guided world modeling with pose-conditioned retrieval.

\paragraph{Implementation details.}
We use Wan2.2-Fun-5B-Control-Camera~\cite{wan2025wan} as the base DiT, with hidden dimension $D=3072$, and retain its UMT5 text cross-attention and Pl\"ucker-ray camera adapter.
The video VAE and text encoder remain frozen.
Rank-32 LoRA adapters are trained in the attention and feed-forward layers, together with the patch-embedding columns appended for the two visual flows, 
while gradients on the original input columns are masked out.
We pre-encode the text, video, and image latents together with \emph{Action flow} and \emph{Object flow}; 
caching these latents removes the encoders from the training loop, which shortens training and lowers VRAM use.
Concatenating these two $48$-channel latents and the one-channel coverage mask with the native input yields $197$ channels.
On Sekai-game-drone, transform-aware pretraining runs for $300$ steps, using RGB frames whose marker pixels are replaced by local transformation axes.
RGB training then continues for $5$ epochs from those weights, with the optimizer reset and the target returned to unmarked video.
DuoWM w/o pretrain skips the $300$-step stage.
The game-domain run uses a learning rate of $10^{-4}$ and a global batch size of $16$.
Real-domain models are trained for two RGB epochs.
Inference generates the $121$-frame clips with $50$ denoising steps, guidance scale $1$, and seed $1$.

\begin{table*}[ht]
\centering
\caption{\textbf{Camera-controlled video generation on Sekai-game-drone.} All methods are evaluated on the same 474 test videos.
All methods undergo $5$ epochs of RGB training on the training split, except the Wan2.2-5B base;
DuoWM additionally uses 300 pretraining steps, whereas DuoWM w/o pretrain skips this stage.
The best, second-best, and third-best results in each metric
are highlighted in blue, yellow, and purple, respectively;
the best results are also shown in bold.}
\label{tab:main}
\small
\setlength{\tabcolsep}{4pt}
\renewcommand{\arraystretch}{1.15}
\resizebox{0.8\textwidth}{!}{%
\begin{tabular}{lcccccccc}
\toprule
\textbf{Model} & \textbf{Year} & \textbf{Venue}
& \multicolumn{2}{c}{\textbf{Video Quality}}
& \multicolumn{1}{c}{\textbf{3D Consistency}}
& \multicolumn{3}{c}{\textbf{Trajectory Following}} \\
\cmidrule(lr){4-5}
\cmidrule(lr){6-6}
\cmidrule(lr){7-9}
& & &
FVD $\downarrow$ &
LPIPS $\downarrow$ &
MEt3R $\downarrow$ &
AUC@30 $\uparrow$ &
AUC@15 $\uparrow$ &
CosSim $\uparrow$ \\
\midrule
MotionCtrl
& 2024 & SIGGRAPH
& 200.8430 & 0.3560 & 0.3583
& 0.2402 & 0.0956 & 0.2599 \\

Wan2.2-5B base
& 2025 & --
& 298.3193 & 0.3054 & 0.3121
& 0.3763 & 0.1699 & 0.2841 \\

Wan2.2-Control-Camera
& 2025 & --
& 75.6100 & 0.2711 & 0.3106
& \rankthird{0.5116}
& \rankthird{0.3232}
& 0.3199 \\

Real-CamI2V
& 2025 & ICCV
& 139.1300 & 0.3148 & 0.3733
& 0.3303 & 0.1643 & 0.2042 \\

CameraCtrl
& 2025 & ICLR
& 221.8947 & 0.3254 & 0.3691
& 0.3333 & 0.1639 & 0.2841 \\

Captain Safari
& 2026 & CVPR
& \ranksecond{69.8325}
& \rankthird{0.2606}
& \ranksecond{0.3000}
& 0.5003
& 0.3088
& \rankthird{0.3325} \\
\midrule
DuoWM
& 2026 & --
& \rankfirst{65.9881}
& \rankfirst{0.2593}
& \rankfirst{0.2994}
& \rankfirst{0.5719}
& \rankfirst{0.3747}
& \rankfirst{0.3595} \\

DuoWM w/o pretrain
& 2026 & --
& \rankthird{70.3296}
& \ranksecond{0.2596}
& \rankthird{0.3017}
& \ranksecond{0.5537}
& \ranksecond{0.3522}
& \ranksecond{0.3580} \\
\bottomrule
\end{tabular}%
}
\end{table*}

\begin{table*}[ht]
\centering
\caption{\textbf{Benchmark camera-controlled video generation on Sekai-real-drone.} All methods are evaluated on the 402 test videos.
All methods undergo $2$ epochs of RGB training on the training split, except the Wan2.2-5B base.
DuoWM additionally uses pretraining, whereas DuoWM w/o pretrain skips this stage.}
\label{tab:real_results}
\small
\setlength{\tabcolsep}{4pt}
\renewcommand{\arraystretch}{1.15}
\resizebox{0.8\textwidth}{!}{%
\begin{tabular}{lcccccccc}
\toprule
\textbf{Model} & \textbf{Year} & \textbf{Venue}
& \multicolumn{2}{c}{\textbf{Video Quality}}
& \multicolumn{1}{c}{\textbf{3D Consistency}}
& \multicolumn{3}{c}{\textbf{Trajectory Following}} \\
\cmidrule(lr){4-5}
\cmidrule(lr){6-6}
\cmidrule(lr){7-9}
& & & FVD $\downarrow$ & LPIPS $\downarrow$ & MEt3R $\downarrow$
& AUC@30 $\uparrow$ & AUC@15 $\uparrow$ & CosSim $\uparrow$ \\
\midrule
Wan2.2-5B base & 2025 & --
& 299.6678 & 0.1843 & 0.1175 & 0.2173 & 0.1147 & 0.2655 \\
Wan2.2-Control-Camera & 2025 & --
& \rankthird{213.0338} & \rankthird{0.1422} & \ranksecond{0.1084}
& 0.4856 & 0.3125 & \rankthird{0.2949} \\
Captain Safari & 2026 & CVPR
& 231.6737 & 0.1594 & 0.1255
& \rankthird{0.5218} & \ranksecond{0.3521} & 0.2675 \\
\midrule
DuoWM & -- & --
& \rankfirst{191.7696} & \rankfirst{0.1250} & \rankfirst{0.1044}
& \rankfirst{0.5409} & \rankfirst{0.3709} & \rankfirst{0.3110} \\
DuoWM w/o pretrain & -- & --
& \ranksecond{202.0693} & \ranksecond{0.1267} & \rankthird{0.1086}
& \ranksecond{0.5248} & \rankthird{0.3508} & \ranksecond{0.3014} \\
\bottomrule
\end{tabular}%
}
\end{table*}

\begin{table*}[t]
\centering
\caption{\textbf{Ablation studies on Sekai-game-drone.} All variants are evaluated on the same 474 test videos. Left: component ablations with five RGB training epochs; full DuoWM additionally uses 300 pretraining steps. Right: warping and pretraining comparisons with one RGB training epoch; the three pretrained variants use 60 pretraining steps. Variants marked w/o PT omit pretraining. Each within-group comparison changes one factor, keeping other settings fixed. PT: pretraining; OF/AF: Object/Action Flow; WW: WorldWarp-based construction with 500 Gaussian-fitting steps per video; TP/RP: trajectory-/RGB-aware pretraining. Best and second-best results within each block are bold and underlined, respectively.}
\label{tab:ablation}

\scriptsize
\setlength{\tabcolsep}{2pt}
\renewcommand{\arraystretch}{1.08}

\noindent
\begin{minipage}[t]{0.49\textwidth}
\vspace{0pt}
\centering
\renewcommand{\arraystretch}{1.75}
\resizebox{\linewidth}{!}{%
\begin{tabular}{lcccccc}
\toprule
\textbf{Variant}
& \multicolumn{2}{c}{\textbf{Video Quality}}
& \textbf{3D Cons.}
& \multicolumn{3}{c}{\textbf{Trajectory Following}} \\
\cmidrule(lr){2-3}
\cmidrule(lr){4-4}
\cmidrule(lr){5-7}
& FVD $\downarrow$ & LPIPS $\downarrow$ & MEt3R $\downarrow$
& AUC@30 $\uparrow$ & AUC@15 $\uparrow$ & CosSim $\uparrow$ \\
\midrule
\rowcolor{black!8}
\multicolumn{7}{l}{\textit{Component Ablation}} \\
DuoWM
& \textbf{65.9881} & \textbf{0.2593} & \textbf{0.2994}
& \textbf{0.5719} & \textbf{0.3747} & \textbf{0.3595} \\
w/o PT
& \underline{70.3296} & \underline{0.2596} & \underline{0.3017}
& \underline{0.5537} & \underline{0.3522} & \underline{0.3580} \\
w/o PT, w/o OF
& 76.0832 & 0.2642 & 0.3028
& 0.5324 & 0.3361 & 0.3550 \\
w/o PT, w/o OF, w/o AF
& 75.6100 & 0.2711 & 0.3106
& 0.5116 & 0.3232 & 0.3199 \\
\bottomrule
\end{tabular}%
}
\end{minipage}
\hfill
\begin{minipage}[t]{0.49\textwidth}
\vspace{0pt}
\centering
\renewcommand{\arraystretch}{1.08}
\resizebox{\linewidth}{!}{%
\begin{tabular}{lcccccc}
\toprule
\textbf{Variant}
& \multicolumn{2}{c}{\textbf{Video Quality}}
& \textbf{3D Cons.}
& \multicolumn{3}{c}{\textbf{Trajectory Following}} \\
\cmidrule(lr){2-3}
\cmidrule(lr){4-4}
\cmidrule(lr){5-7}
& FVD $\downarrow$ & LPIPS $\downarrow$ & MEt3R $\downarrow$
& AUC@30 $\uparrow$ & AUC@15 $\uparrow$ & CosSim $\uparrow$ \\
\midrule
\rowcolor{black!8}
\multicolumn{7}{l}{\textit{Warping strategy}} \\
DuoWM
& \textbf{89.5135} & \textbf{0.2828} & \textbf{0.3197}
& \textbf{0.4720} & \textbf{0.2547} & \textbf{0.3240} \\
WW
& 94.1659 & 0.2867 & 0.3217
& 0.4190 & 0.2308 & 0.3181 \\
\midrule
\rowcolor{black!8}
\multicolumn{7}{l}{\textit{Pretraining Strategy}} \\
DuoWM
& \textbf{89.5135} & \underline{0.2828} & \textbf{0.3197}
& \textbf{0.4720} & \underline{0.2547} & \underline{0.3240} \\
DuoWM w/o PT
& \underline{91.3989} & \textbf{0.2826} & \underline{0.3198}
& \underline{0.4668} & \textbf{0.2572} & \textbf{0.3314} \\
TP
& 103.7779 & 0.2869 & 0.3224
& 0.4131 & 0.2014 & 0.3062 \\
RP
& 96.6638 & 0.2845 & 0.3201
& 0.4644 & 0.2462 & 0.3187 \\
\bottomrule
\end{tabular}%
}
\end{minipage}

\end{table*}

\begin{figure}[t]
    \centering
\includegraphics[width=0.832\textwidth,height=0.388\textheight,keepaspectratio]{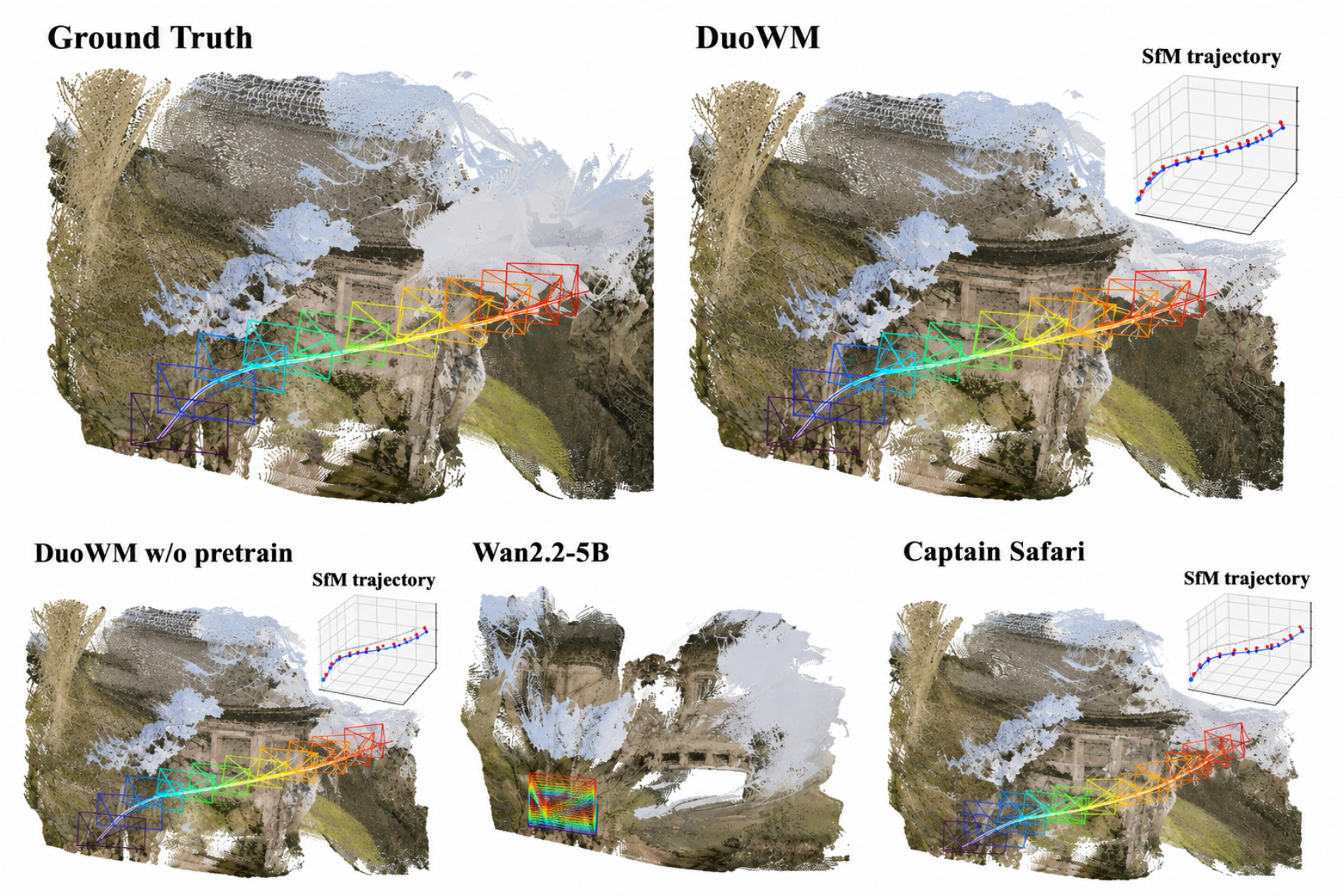}
    \caption{\textbf{Qualitative comparison of 3D reconstructions and camera trajectories.} Ground truth is compared with DuoWM, DuoWM without pretraining, Wan2.2-5B, and Captain Safari. In the SfM insets, gray lines denote ground-truth trajectories, red points denote reconstructed relative camera positions, and blue lines connect these positions.}
    \label{fig:3d_construction}
\end{figure}

\subsection{Camera-Controlled Video Generation}
\paragraph{Video quality}
As shown in Table~\ref{tab:main}, DuoWM attains the best video quality on Sekai-game-drone.
Relative to Wan2.2-Control-Camera, it achieves a substantially lower FVD (65.9881 vs.\ 75.6100), a 12.7\% reduction, and a lower LPIPS (0.2593 vs.\ 0.2711).
Compared with Captain Safari, the strongest prior model on these two metrics, DuoWM further lowers FVD (65.9881 vs.\ 69.8325) and slightly improves LPIPS (0.2593 vs.\ 0.2606), 
indicating more stable temporal dynamics and sharper frames.
The same ordering holds on Sekai-real-drone (Table~\ref{tab:real_results}), 
where DuoWM lowers FVD (191.7696 vs.\ 231.6737) and LPIPS (0.1250 vs.\ 0.1594) relative to Captain Safari, reductions of 17.2\% and 21.6\%.
DuoWM w/o pretrain also lowers FVD on Sekai-real-drone (202.0693 vs.\ 231.6737), but on Sekai-game-drone its FVD is slightly higher than Captain Safari (70.3296 vs.\ 69.8325).

\paragraph{3D consistency.}
DuoWM also gives the lowest MEt3R on both benchmarks.
On Sekai-game-drone, its score of 0.2994 is slightly lower than Captain Safari (0.3000) and Wan2.2-Control-Camera (0.3106).
On Sekai-real-drone, MEt3R falls to 0.1044, below Wan2.2-Control-Camera (0.1084) and Captain Safari (0.1255), indicating more consistent cross-view geometry.
Without pretraining, MEt3R remains lower than Captain Safari on Sekai-real-drone (0.1086 vs.\ 0.1255) but is slightly higher on Sekai-game-drone (0.3017 vs.\ 0.3000).

\begin{figure*}[t]
    \centering
    \includegraphics[width=0.98\textwidth,height=0.46\textheight,keepaspectratio]{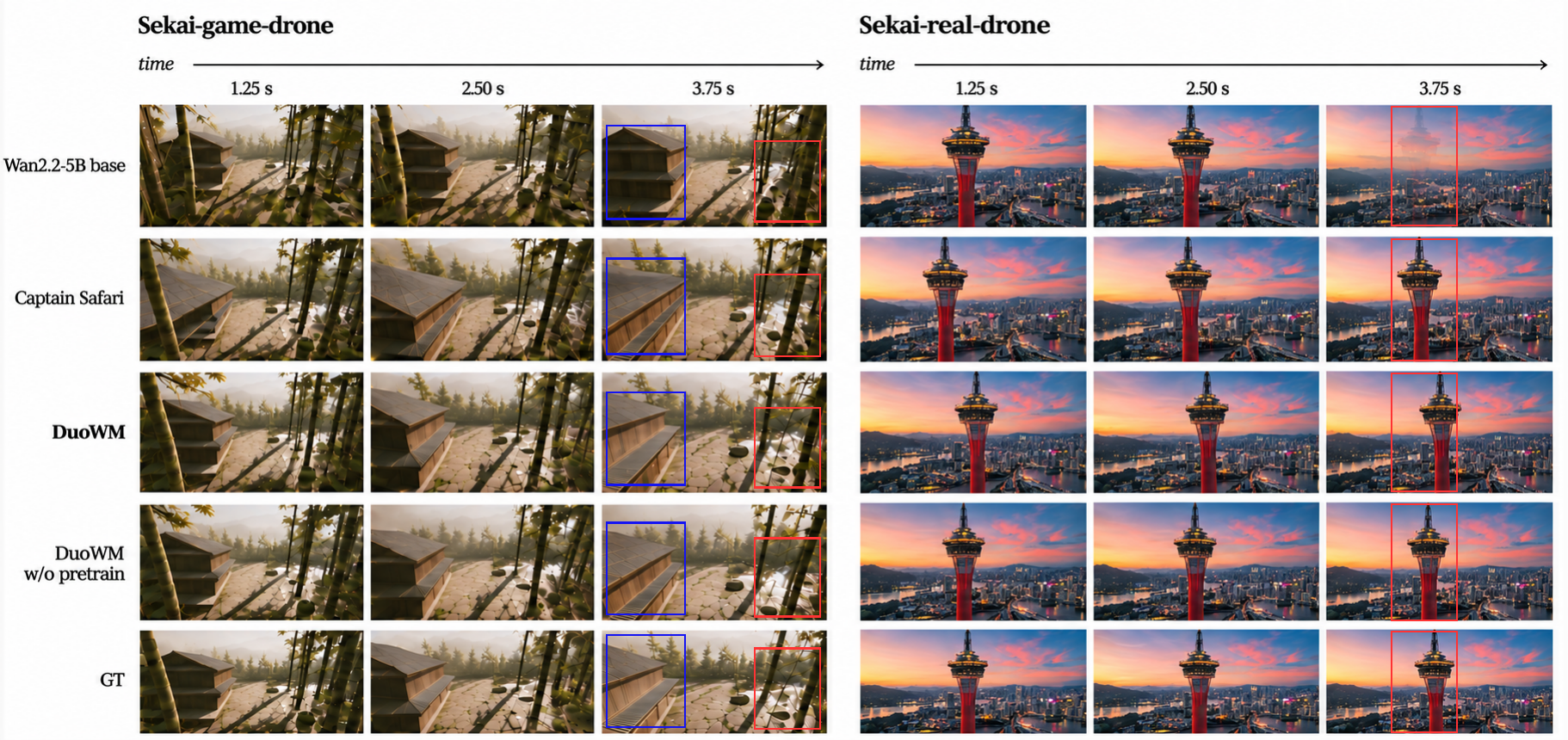}
    \caption{\textbf{DuoWM keeps the commanded viewpoint and scene structure.}
    Qualitative results on Sekai-game-drone and Sekai-real-drone at 1.25\,s, 2.50\,s, and 3.75\,s.
    Rows compare Wan2.2-5B base, Captain Safari, DuoWM, DuoWM without pretraining, and the ground truth.}
    \label{fig:method_comparison}
\end{figure*}

\paragraph{Trajectory following.}
DuoWM delivers the most accurate trajectory following on both benchmarks.
As shown in Table~\ref{tab:main}, on Sekai-game-drone it achieves the highest AUC@30 (0.5719) and AUC@15 (0.3747), along with the best cosine similarity (0.3595), 
exceeding Wan2.2-Control-Camera (0.5116, 0.3232, and 0.3199) and Captain Safari (0.5003, 0.3088, and 0.3325).
On Sekai-real-drone it again leads, with AUC@30 of 0.5409, AUC@15 of 0.3709, and cosine similarity of 0.3110.
Figure.~\ref{fig:method_comparison} shows the corresponding visual advantage: the generated views stay on the commanded viewpoint instead of drifting toward a nearby but incorrect direction.
Figure.~\ref{fig:3d_construction} makes the same point in the recovered geometry, where the SfM trajectories of DuoWM remain closer to the ground-truth path than those of Wan2.2-5B and Captain Safari.
Without pretraining, both AUC thresholds on Sekai-game-drone still exceed Captain Safari (0.5537 vs.\ 0.5003 and 0.3522 vs.\ 0.3088).
On Sekai-real-drone, AUC@30 remains slightly higher (0.5248 vs.\ 0.5218), while AUC@15 falls just short (0.3508 vs.\ 0.3521).
With the same first-frame-and-trajectory \emph{Object flow}, the gains in visual fidelity, geometry, and camera control appear on both datasets.

\subsection{Ablation Studies}
Table~\ref{tab:ablation} evaluates individual design choices. Component ablations use five RGB training epochs, with 300 pretraining steps for full DuoWM. Warping and pretraining-strategy comparisons use one RGB epoch and, when enabled, 60 pretraining steps. Real-domain comparisons use two RGB epochs.

\paragraph{Action Flow and Object Flow.}
The two visual flows improve appearance and camera control together (Table~\ref{tab:ablation}).
Adding \emph{Action flow} to Wan2.2-Control-Camera, still without pretraining, raises AUC@30 by 4.1\% (0.5324 vs.\ 0.5116) and CosSim by 11.0\% (0.3550 vs.\ 0.3199), 
and lowers LPIPS and MEt3R by 2.5\%, with only a 0.6\% increase in FVD (76.0832 vs.\ 75.6100).
Adding \emph{Object flow} then improves all six metrics. FVD falls from 76.0832 to 70.3296, a 7.6\% reduction, and AUC@30 and AUC@15 rise from 0.5324 and 0.3361 to 0.5537 and 0.3522.
\emph{Action flow} places the camera command in each frame, while \emph{Object flow} binds that motion to the observed surfaces. 
Removing either one leaves part of this geometric prior unused.

\paragraph{Effect of transform-aware pretraining.}
At the full training budget, pretraining improves all six metrics on both datasets.
On Sekai-game-drone it lowers FVD from 70.3296 to 65.9881, a 6.2\% reduction, and MEt3R from 0.3017 to 0.2994, with gains in LPIPS and all three trajectory metrics.
On Sekai-real-drone, FVD decreases by 5.1\% (191.7696 vs.\ 202.0693) and MEt3R by 3.9\% (0.1044 vs.\ 0.1086), while AUC@30, AUC@15, and CosSim improve by 3.1\%, 5.7\%, and 3.2\%.
Figure.~\ref{fig:method_comparison} and Figure.~\ref{fig:3d_construction} show the same effect. With pretraining, the cabin and the tower stay aligned with the ground truth. 
Without it, the viewpoint and the recovered trajectory drift.
The shorter budget of $60$ pretraining steps and one RGB epoch has not yet resolved this trade-off. 
Removing pretraining slightly improves LPIPS (0.2826 vs.\ 0.2828), AUC@15 (0.2572 vs.\ 0.2547), and CosSim (0.3314 vs.\ 0.3240), 
but raises FVD from 89.5135 to 91.3989 and lowers AUC@30 from 0.4720 to 0.4668.
Only the full schedule turns the geometric prior into gains on every reported metric.

\paragraph{Object Flow construction.}
Replacing forward warping with WorldWarp degrades all six metrics.
WorldWarp fits a Gaussian scene for $500$ steps per video and renders the object maps from that fit.
After one RGB epoch, forward warping lowers FVD from 94.1659 to 89.5135, a 4.9\% reduction, and raises AUC@30 from 0.4190 to 0.4720 and AUC@15 from 0.2308 to 0.2547.
The initial-frame warp therefore supplies a stronger geometric condition, without a separate per-video Gaussian optimization.

\paragraph{Choice of pretraining representation.}
The pretraining target should carry the same geometry as the \emph{Object flow} condition.
Among local transformation axes, trajectories, and RGB appearance, the axis formulation is best on all six metrics.
Its FVD of 89.5135 is 7.4\% lower than RGB-aware pretraining (96.6638) and 13.7\% lower than trajectory-aware pretraining (103.7779), 
with corresponding gains in MEt3R and trajectory following.
Explicit axes give the denoiser a visible link between surface appearance and rigid motion. Trajectory traces and RGB targets leave that link weaker.

\section{Conclusion}
\label{sec:conclusion}
We introduced DuoWM, a first-person drone world model that generates camera-controlled video from an initial frame and a target trajectory.
It is guided by two complementary visual flows, \emph{Action flow} and \emph{Object flow}, with transform-aware pretraining that ties these geometric cues to appearance.
On both game and real drone benchmarks, DuoWM improves video quality, 3D consistency, and trajectory following over prior camera-controlled world models.
Future work will explore longer-horizon autoregressive generation and closed-loop interaction with flight policies.
We hope DuoWM encourages further research on explicit 6-DoF guidance for persistent, controllable first-person drone world models.

\subsection*{Acknowledgements}
The authors would like to acknowledge the support provided by the Beijing Key Laboratory of UAV Integrated Intelligent Control Advanced Technology. 

\bibliographystyle{unsrt}  
\bibliography{references}  

\clearpage
\setcounter{page}{1}

\section{Additional Implementation Details}
\label{sec:supp_implementation}

\paragraph{Data and training schedules.}
We use 5,118 training and 474 test videos from
Sekai-game-drone~\cite{li2025sekai}.
For the real-domain experiments, we select a subset of
4,368 training and 402 test videos as our real-domain dataset
from the Sekai-real-drone dataset.
All generated clips comprise 121 frames at a resolution
of $1280\times704$ and 24 FPS.
DuoWM undergoes 300 steps of transform-aware pretraining
followed by 5 epochs of RGB training on the game domain,
and receives 2 epochs of RGB training on the real domain.
The supplementary ablations are conducted on Sekai-game-drone
using 60 pretraining steps followed by one RGB training epoch
and are evaluated on the same 474-video test set.

\paragraph{Optimization and sampling.}
All main experiments are conducted on 8 NVIDIA A100 GPUs, each with 80\,GB of memory. 
DuoWM retains the text-conditioning and Pl\"ucker-ray pathways of Wan2.2-Fun-5B-Control-Camera while keeping the video VAE, text encoder, camera adapter, 
and original backbone parameters frozen. 
We optimize only the rank-32 LoRA adapters in the attention and feed-forward layers, together with the newly added patch-embedding columns. 
Each visual flow contributes a 48-channel latent; together with the coverage mask and native inputs, these conditions expand the model input to 197 channels. 
All encoder outputs are cached before training. The main game-domain run uses a learning rate of $10^{-4}$ and a global batch size of 16. 
The RGB-training stage is initialized from the transform-aware pretrained weights, with the optimizer reinitialized. 
At inference, we use 50 denoising steps, a guidance scale of 1, and seed 1.

\begin{figure*}[ht]
\centering
\includegraphics[
    width=\textwidth,
    height=0.38\textheight,
    keepaspectratio,
    draft=false
]{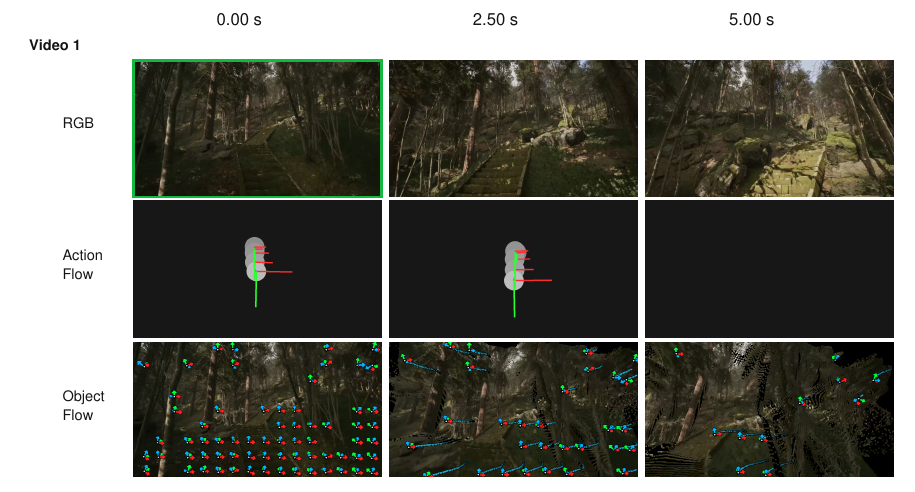}
\caption{\textbf{RGB, \emph{action flow}, and \emph{object flow} examples.}
Each group shows reference RGB frames and corresponding visual
conditions from one Sekai-game-drone video at 0.00, 2.50,
and 5.00 seconds. Green outlines mark the starting RGB frames.
The final \emph{action flow} map is empty because no future camera
pose remains.}
\label{fig:supp_flow_examples}
\end{figure*}

\paragraph{Information available to the generator.}
Across both domains, DuoWM constructs \emph{action flow} and \emph{object flow} from the initial image and the supplied camera trajectory; 
representative conditions are shown in Figure.~\ref{fig:supp_flow_examples}. 
To construct \emph{object flow}, DuoWM first forward-warps the initial appearance and then applies frozen MoSE3 to estimate local geometric cues from the warped sequence. 
The depth scale is calibrated using training videos and then held fixed at inference; no test-video correspondences are used. 
Full training clips are used only to construct supervision, including the geometry-annotated pretraining targets, and are never supplied to the conditioning branches. 
Consequently, the local cues encode correspondences inferred from the initial observation under the requested camera motion, 
rather than measurements of future independent object motion.

\paragraph{Pretraining representations.}
The pretraining ablation compares three paired target-condition representations. 
Transform-aware pretraining overlays local transformation axes on RGB appearance, trajectory-aware pretraining uses positional traces without orientation axes, 
and RGB-aware pretraining uses unmarked RGB appearance. 
Every pretraining experiment consists of two stages: a pretraining stage followed by an RGB-training stage. 
For each variant, the same representation is used in both the reconstruction target and the corresponding \emph{object flow} condition, 
thereby testing the paired representation rather than the reconstruction target alone. 
In the default transform-aware configuration, \emph{object flow} contains warped appearance and local axes during the pretraining stage; 
the retained local axes, added positional traces, and warped appearance together form the \emph{object flow} condition for the subsequent RGB training stage and inference. 
Across all variants, the reconstruction target in the RGB-training stage is unmarked RGB video. 
Because of limited computational resources, the short-budget comparison reported in the main paper trains each pretrained variant for 60 pretraining steps followed by one RGB epoch, 
whereas the variant without pretraining receives only the RGB epoch.

\paragraph{Evaluation and interpretation.}
We follow the same evaluation procedure as the main paper. 
FVD measures distributional similarity between generated and reference videos, and LPIPS measures perceptual differences from the reference frames. 
Following the evaluation scripts of Captain Safari~\cite{chou2026captain}, we compute MEt3R between each generated frame and the ground-truth frame at the same timestamp. 
We interpret this score as reference-based geometric agreement; it does not directly measure consistency across different frames of a generated video. 
For trajectory evaluation, camera poses are recovered with HLoc~\cite{sarlin2019coarse} and COLMAP~\cite{schonberger2016structure}. 
AUC@15 and AUC@30 summarize relative rotation and translation-direction errors at $15^\circ$ and $30^\circ$, respectively. 
CosSim compares flattened camera extrinsics after first-frame canonicalization, without Sim(3) alignment. 
Videos whose poses cannot be recovered are excluded from the trajectory averages, which therefore describe only the successfully reconstructed subset. 

\begin{table*}[t]
\centering
\caption{\textbf{Components of \emph{object flow}.} All variants retain \emph{action flow} and the same coverage mask, and use 60 pretraining steps followed by one RGB epoch on Sekai-game-drone. The best and second-best results in each column are bold and underlined, respectively.}
\label{tab:object}
\small
\setlength{\tabcolsep}{6pt}
\resizebox{\textwidth}{!}{%
\begin{tabular}{lrrrrrr}
\toprule
Variant & FVD $\downarrow$ & LPIPS $\downarrow$ & MEt3R $\downarrow$ & AUC@30 $\uparrow$ & AUC@15 $\uparrow$ & CosSim $\uparrow$ \\
\midrule
Warp only & 94.0281 & 0.2960 & 0.3392 & 0.4497 & 0.2356 & 0.3092 \\
Warp + local axes & 92.7087 & \underline{0.2835} & 0.3237 & 0.4654 & 0.2432 & 0.3164 \\
Warp + traces & \underline{91.3831} & 0.2893 & \underline{0.3218} & \textbf{0.4733} & \underline{0.2480} & \underline{0.3178} \\
Warp + axes + traces & \textbf{89.5135} & \textbf{0.2828} & \textbf{0.3197} & \underline{0.4720} & \textbf{0.2547} & \textbf{0.3240} \\
Axes only, without RGB warp & 101.1071 & 0.3025 & 0.3373 & 0.4464 & 0.2285 & 0.3024 \\
\bottomrule
\end{tabular}}
\end{table*}

\begin{table*}[t]
\centering
\caption{\textbf{Orientation and depth cues in \emph{action flow}.} All configurations use five requested pose markers and the same short-budget training schedule. ``Depth'' denotes depth-dependent marker brightness. The best and second-best results in each column are bold and underlined, respectively.}
\label{tab:action}
\small
\setlength{\tabcolsep}{4pt}
\resizebox{0.9\textwidth}{!}{%
\begin{tabular}{lccrrrrrr}
\toprule
Configuration & Axes & Depth
& FVD $\downarrow$ & LPIPS $\downarrow$ & MEt3R $\downarrow$
& AUC@30 $\uparrow$ & AUC@15 $\uparrow$ & CosSim $\uparrow$ \\
\midrule
Position markers + axes & Yes & No
& 91.7185 & \underline{0.29099} & \underline{0.32080}
& \underline{0.4669} & \underline{0.2459} & \underline{0.3096} \\

Position markers + depth & No & Yes
& \textbf{88.6680} & 0.29524 & 0.32850
& 0.4362 & 0.2144 & 0.3006 \\

Complete Action Flow & Yes & Yes
& \underline{89.5135} & \textbf{0.28280} & \textbf{0.31970}
& \textbf{0.4720} & \textbf{0.2547} & \textbf{0.3240} \\
\bottomrule
\end{tabular}}
\end{table*}

\begin{table*}[t]
\centering
\caption{\textbf{Number of requested \emph{action flow} markers.} All variants use 60 pretraining steps and one RGB epoch. 
Counts denote the number of future camera-trajectory points drawn in \emph{action flow}, before visibility filtering. The best and second-best results in each column are bold and underlined, respectively.}
\label{tab:count}
\small
\setlength{\tabcolsep}{6pt}
\resizebox{0.85\textwidth}{!}{%
\begin{tabular}{lrrrrrr}
\toprule
Requested markers
& FVD $\downarrow$ & LPIPS $\downarrow$ & MEt3R $\downarrow$
& AUC@30 $\uparrow$ & AUC@15 $\uparrow$ & CosSim $\uparrow$ \\
\midrule
1
& 90.9860 & \underline{0.28537} & 0.32673
& 0.4661 & 0.2464 & 0.3093 \\

3
& \underline{89.8170} & 0.29209 & \textbf{0.31931}
& 0.4689 & 0.2485 & \textbf{0.3270} \\

5 (default)
& \textbf{89.5135} & \textbf{0.28280} & \underline{0.31970}
& \textbf{0.4720} & \textbf{0.2547} & \underline{0.3240} \\

7
& 91.3680 & 0.28760 & 0.32041
& \underline{0.4696} & \underline{0.2505} & 0.3139 \\
\bottomrule
\end{tabular}}
\end{table*}

\section{Supplementary Ablation Studies}
\label{sec:supp_ablation}
These experiments examine the information encoded by \emph{object flow}, the orientation and depth cues in \emph{action flow}, 
and the number of future camera-trajectory points drawn in \emph{action flow}. 
Within each comparison, only the indicated design factor is varied, while all other training and sampling settings remain fixed. 
The complete configuration uses the same default setting in all three tables. 

\subsection{Components of Object Flow}
\label{sec:supp_object}
The third row of Figure.~\ref{fig:supp_flow_examples} shows the \emph{object flow} conditions, aligned with the reference RGB frames above. 
We compare five \emph{object flow} variants: warped appearance alone, warped appearance with local axes, warped appearance with traces, the complete representation, 
and local axes without warped appearance. 
\emph{Action flow} is held fixed. 
All variants share the same source warp, coverage mask, anchors, visibility rules, and input capacity. 
The axes-only variant renders the same projected axes on a dark background while retaining the coverage mask.

Table~\ref{tab:object} shows that adding local axes to warped appearance improves all six metrics, lowering FVD from 94.0281 to 92.7087 and raising AUC@30 from 0.4497 to 0.4654. 
Adding traces instead of axes also improves all six metrics over warp only and gives the highest AUC@30, 0.4733. 
The complete representation, combining the warp, local axes, and traces, achieves the best FVD, LPIPS, MEt3R, AUC@15, and CosSim. 
Its FVD of 89.5135 is 4.8\% lower than that of warp only, while its AUC@30 of 0.4720 is just below the trace-only addition. 
Removing the warped appearance and retaining only the projected axes lowers every metric relative to warp plus axes, 
so the geometric markers are most effective together with the observed surface appearance.

\subsection{Orientation and Depth Cues in Action Flow}
\label{sec:supp_action}
The second row of Figure.~\ref{fig:supp_flow_examples} shows the \emph{action flow} conditions, aligned with the reference RGB frames above. 
We compare three configurations that each draw five future camera-trajectory points: orientation axes with constant marker brightness, depth-dependent brightness without axes, 
and the complete representation containing both. 
The projected centers, marker radius, \emph{object flow}, and current-camera reference update remain fixed.

Table~\ref{tab:action} shows that adding depth-dependent brightness to the axis-bearing markers improves all six metrics. 
Starting from depth-dependent markers, adding orientation axes improves LPIPS, MEt3R, and all three trajectory metrics: AUC@30 rises from 0.4362 to 0.4720, 
and AUC@15 rises from 0.2144 to 0.2547. 
The complete representation achieves the best LPIPS, MEt3R, AUC@30, AUC@15, and CosSim. 
Its FVD of 89.5135 is slightly above the 88.6680 obtained by depth-dependent markers without axes. 
Orientation axes and depth-dependent brightness are therefore complementary: together they give the strongest perceptual agreement, 
reference-based geometric agreement, and camera-trajectory following among the tested configurations.

\subsection{Number of Requested Pose Markers}
\label{sec:supp_count}

We study how the density of future camera-pose markers affects
\emph{Action Flow}. The requested marker count, denoted by $N$,
takes values in $\{1,3,5,7\}$, while rendering parameters and
training settings remain fixed. All variants retain orientation
axes and depth-dependent brightness.

Let $T$ denote the total number of video frames and $t$ the
current frame index, ranging from $0$ to $T-1$.
We denote the number of remaining future frames by $r_t$:
\begin{equation}
r_t=T-1-t.
\label{eq:marker_remaining}
\end{equation}
If no future frame remains, the Action Flow map contains only
the background and sampling stops.

Otherwise, let $a_t$ and $b_t$ denote the inclusive starting
and ending frame indices of the sampling interval.
The nominal window spans 6--36 frames ahead of the current frame.
Its endpoint is capped at the final video frame; when fewer
than six future frames remain, the interval includes all
remaining future frames:
\begin{equation}
\begin{aligned}
a_t &=
\begin{cases}
t+6, & r_t\geq6,\\
t+1, & 1\leq r_t<6,
\end{cases}\\
b_t &= \min(t+36,T-1).
\end{aligned}
\label{eq:marker_window}
\end{equation}

For multiple markers, we divide this interval into $N-1$
equal segments and sample the segment boundaries, including
both endpoints. For a single marker, we select the temporal
midpoint. Let $k\in\{0,\ldots,N-1\}$ index the requested samples,
and let $j_{t,k}$ denote the corresponding future frame index.
Using $\operatorname{round}$ to select the nearest integer, we obtain
\begin{equation}
j_{t,k}=
\begin{cases}
\operatorname{round}\!\left(\dfrac{a_t+b_t}{2}\right),
& N=1,\\[6pt]
\operatorname{round}\!\left(
a_t+\dfrac{k(b_t-a_t)}{N-1}
\right),
& N>1.
\end{cases}
\label{eq:marker_sampling}
\end{equation}
Duplicate indices are removed after rounding.
If only one future frame remains, all requested samples
reduce to that frame.

We then project the selected camera poses into the current
camera view using its pose and intrinsics.
Following the visibility criteria in
Section~\ref{sec:action_flow}, we retain markers whose centers
have positive depth and lie inside the image, and clip their
axis segments at the image boundary.
Rejected markers are not replaced.
The number of rendered markers can therefore be smaller
than the requested count because of limited remaining frames
or visibility. If no marker survives, the map contains only
the background. This procedure is repeated at every frame,
updating both the sampling interval and the reference viewpoint.

Table~\ref{tab:count} shows that the default of five points achieves the lowest FVD and LPIPS and the highest AUC@30 and AUC@15. 
Three points achieve the lowest MEt3R, 0.31931, and the highest CosSim, 0.3270. 
Increasing the count from five to seven worsens every reported metric. 
Five points therefore give the strongest combination of video quality and angular pose accuracy among the tested settings, 
while three points remain slightly better on reference-based geometric agreement and extrinsic cosine similarity.

\section{Summary}
\label{sec:supp_scope}
The supplementary experiments provide component-level evidence for DuoWM's dual-flow design. 
In \emph{object flow}, adding either local axes or traces improves every metric over warped appearance alone, 
and retaining the warped appearance improves every metric over axes alone. 
In \emph{action flow}, orientation axes and depth-dependent brightness together give the best LPIPS, MEt3R, and trajectory scores among the tested configurations, 
while depth-dependent markers without axes retain a slightly lower FVD. 
The marker-count comparison supports the five-point default, which achieves the lowest FVD and LPIPS and the highest AUC@30 and AUC@15. 
Together with the main benchmark, 
these component results support representing camera commands and local scene geometry as complementary visual conditions for drone video generation.

\section{Examples}
\label{sec:supp_examples}
\begin{figure*}[ht]
\centering
\includegraphics[
    width=\textwidth,
    height=0.98\textheight,
    keepaspectratio,
    draft=false
]{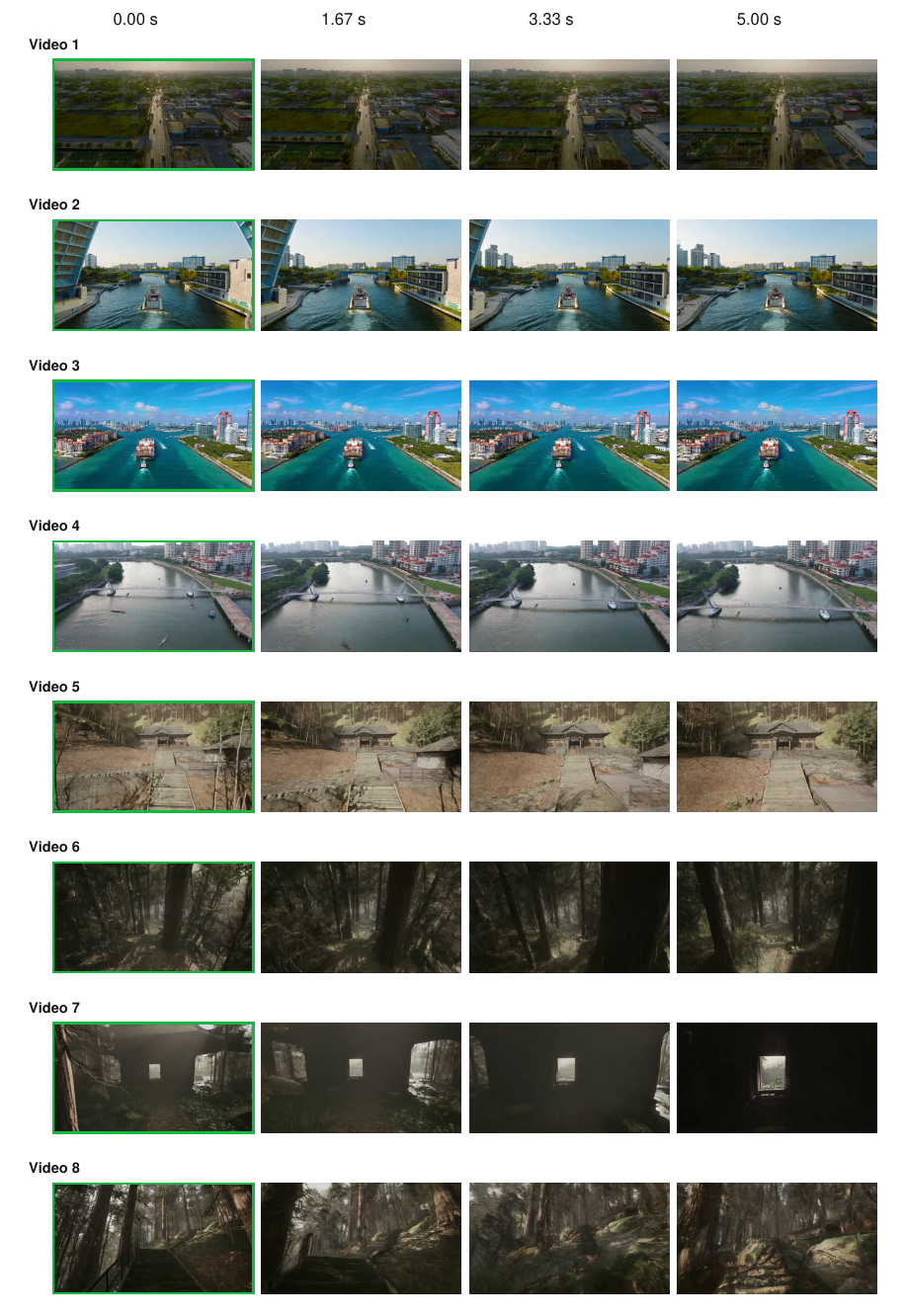}
\caption{\textbf{Generated video examples.}
Videos 1--4 show Sekai-real-drone, and Videos 5--8 show
Sekai-game-drone. Each row presents frames 0, 40, 80, and 120
from one generated video at 24 FPS.
Green outlines mark the starting frames.}
\label{fig:supp_video_examples}
\end{figure*}

Figure~\ref{fig:supp_video_examples} presents generated video examples by our proposed DuoWM, with successive frames illustrating how scene appearance evolves along the supplied camera trajectories. Figure~\ref{fig:supp_flow_examples} shows RGB frames together with the corresponding \emph{action flow} and \emph{object flow} for a single game-domain video. \emph{Action flow} visualizes future camera poses in the current camera view, while \emph{object flow} combines warped appearance with local coordinate axes and traces. Green outlines mark the starting RGB frames; the final \emph{action flow} map is empty because no future camera pose remains.

\end{document}